\documentclass[runningheads]{llncs}
\usepackage{graphicx,amsmath,bm,subfig,threeparttable,xcolor}

\usepackage{cite}

\newcommand{\DCT}{D_{\mbox{\tiny{CT}}}}
\newcommand{\DMR}{D_{\mbox{\tiny{MR}}}}

\newcommand{\GCT}{G_{\mbox{\tiny{CT}}}}
\newcommand{\GMR}{G_{\mbox{\tiny{MR}}}}

\newcommand{\ICT}{I_{\mbox{\tiny{CT}}}}
\newcommand{\IMR}{I_{\mbox{\tiny{MR}}}}

\newcommand{\NCT}{N_{\mbox{\tiny{CT}}}}
\newcommand{\NMR}{N_{\mbox{\tiny{MR}}}}

\newcommand{\KCT}{K_{\mbox{\tiny{CT}}}}
\newcommand{\KMR}{K_{\mbox{\tiny{MR}}}}

\newcommand{\LGAN}{\mathcal{L}_{\mbox{\tiny{GAN}}}}
\newcommand{\Lc}{\mathcal{L}_{\mbox{\tiny{cycle}}}}
\newcommand{\Lstructure}{\mathcal{L}_{\mbox{\tiny{structure}}}}

\begin{document}
\title{Unpaired Brain MR-to-CT Synthesis using a Structure-Constrained CycleGAN}
\titlerunning{Structure-Constrained CycleGAN}
\author{Heran Yang\inst{1,2} \and
	Jian Sun\inst{1} \and
	Aaron Carass\inst{2} \and
	Can Zhao\inst{2} \and
	Junghoon Lee\inst{3} \and
	Zongben Xu\inst{1} \and
	Jerry Prince\inst{2}}
\authorrunning{H. Yang et al.}
\institute{School of Mathematics and Statistics, Xi'an Jiaotong University, China \email{yhr.7017@stu.xjtu.edu.cn}
	\and Department of Electrical and Computer Engineering, Johns Hopkins University, USA
	\and Department of Radiation Oncology and Molecular Radiation Sciences, Johns Hopkins University, USA
}
\maketitle
\begin{abstract}
The cycleGAN is becoming an influential method in medical image synthesis. However, due to a lack of direct constraints between input and synthetic images, the cycleGAN cannot guarantee structural consistency between these two images, and such consistency is of extreme importance in medical imaging. To overcome this, we propose a structure-constrained cycleGAN for brain MR-to-CT synthesis using unpaired data that defines an extra structure-consistency loss based on the modality independent neighborhood descriptor to constrain structural consistency. Additionally, we use a position-based selection strategy for selecting training images instead of a completely random selection scheme. Experimental results on synthesizing CT images from brain MR images demonstrate that our method is better than the conventional cycleGAN and approximates the cycleGAN trained with paired data.
\end{abstract}

\section{Introduction}
\label{sec:introduction}
Magnetic resonance~(MR) imaging has been widely utilized to diagnose patients, as it is
non-ionizing, non-invasive, and has a range of contrast mechanisms.
However, MR images do not directly provide electron density information, which is
essential for some applications such as MR-based radiotherapy
treatment planning or attenuation correction in hybrid PET/MR
scanners. A straightforward solution is to separately scan a computed
tomography~(CT) image, but this is time-consuming, costly, potentially harmful to patients,
and requires accurate MR/CT registrations. Therefore, to
avoid the CT scan, a variety of approaches have been proposed to
synthesize CT images from available MR images~\cite{chartsias2017,
hofmann2011, Roy2017, wolterink2017, zhang2018}.
For example, by using paired MR and CT atlases, atlas-based methods~\cite{hofmann2011} first register multiple atlas MR images to a subject MR image, and then the warped atlas CT images are combined to synthesize a subject CT image.
Deep learning-based methods~\cite{Roy2017} have designed different convolutional neural network~(CNN) structures to directly learn the MR-to-CT mapping.

\begin{figure*}[!tp]
	\setlength{\abovecaptionskip}{2mm}
	\setlength{\belowcaptionskip}{-3mm}
	\centering
	\subfloat[]{
		\label{fig:motivation_gt}
		\begin{minipage}[t]{0.225\textwidth}
			\centering
			\includegraphics[width=2.6cm]{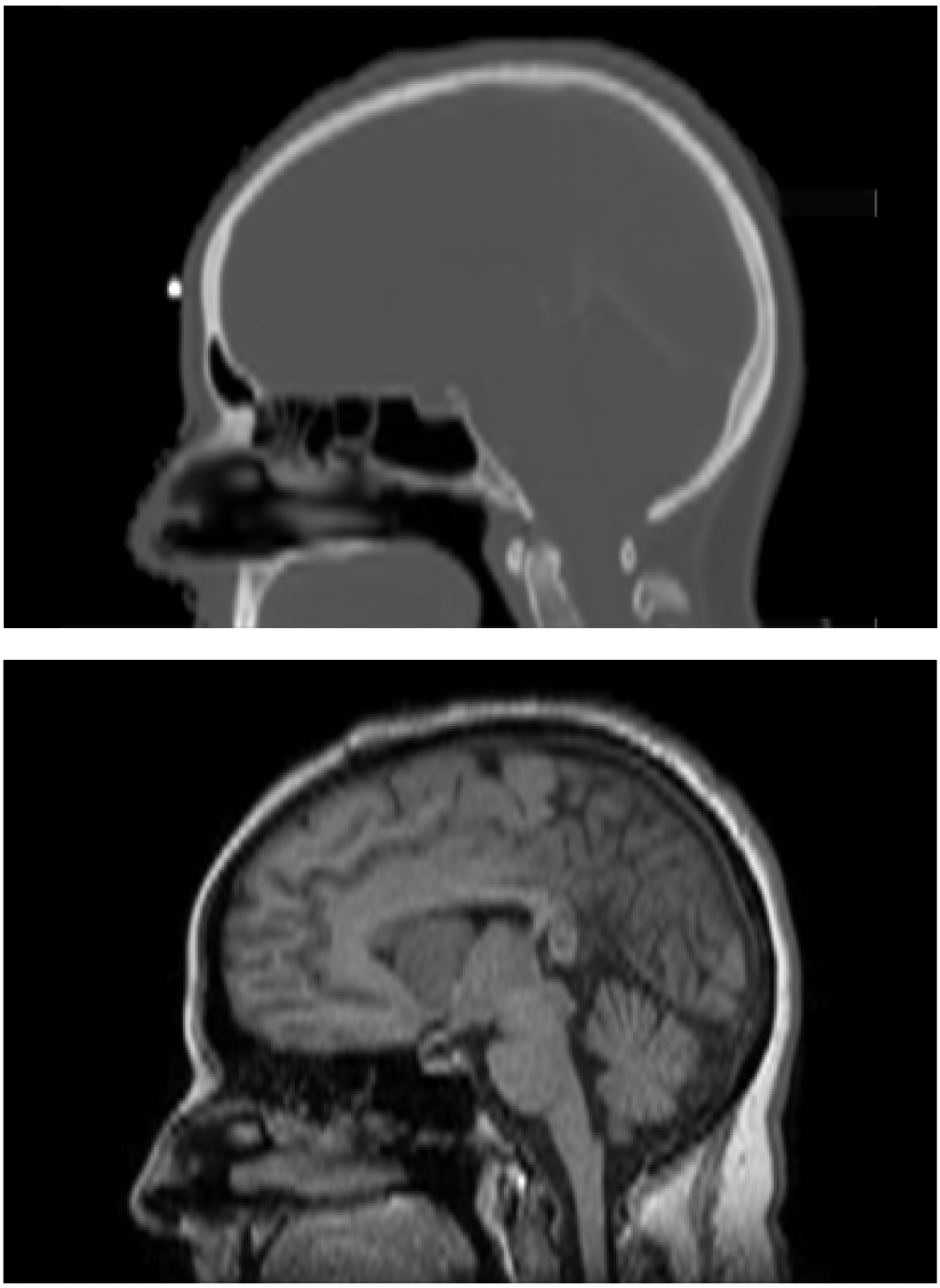}
		\end{minipage}
	}
	\subfloat[]{
		\label{fig:motivation_syn}
		\begin{minipage}[t]{0.225\textwidth}
			\centering
			\includegraphics[width=2.6cm]{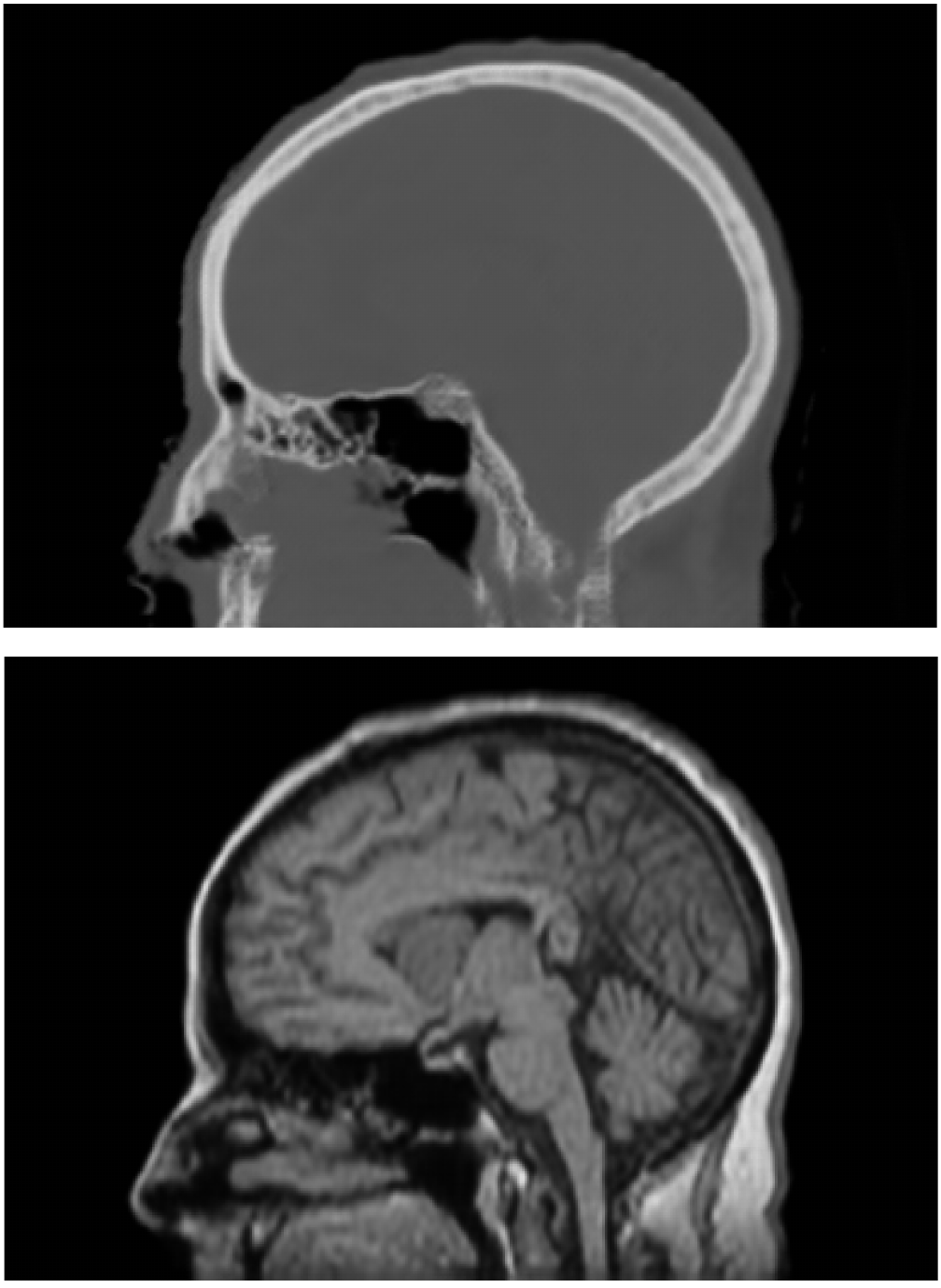}
		\end{minipage}
	}
	\subfloat[]{
		\label{fig:motivation_diff}
		\begin{minipage}[t]{0.24\textwidth}
			\centering
			\includegraphics[width=3.07cm]{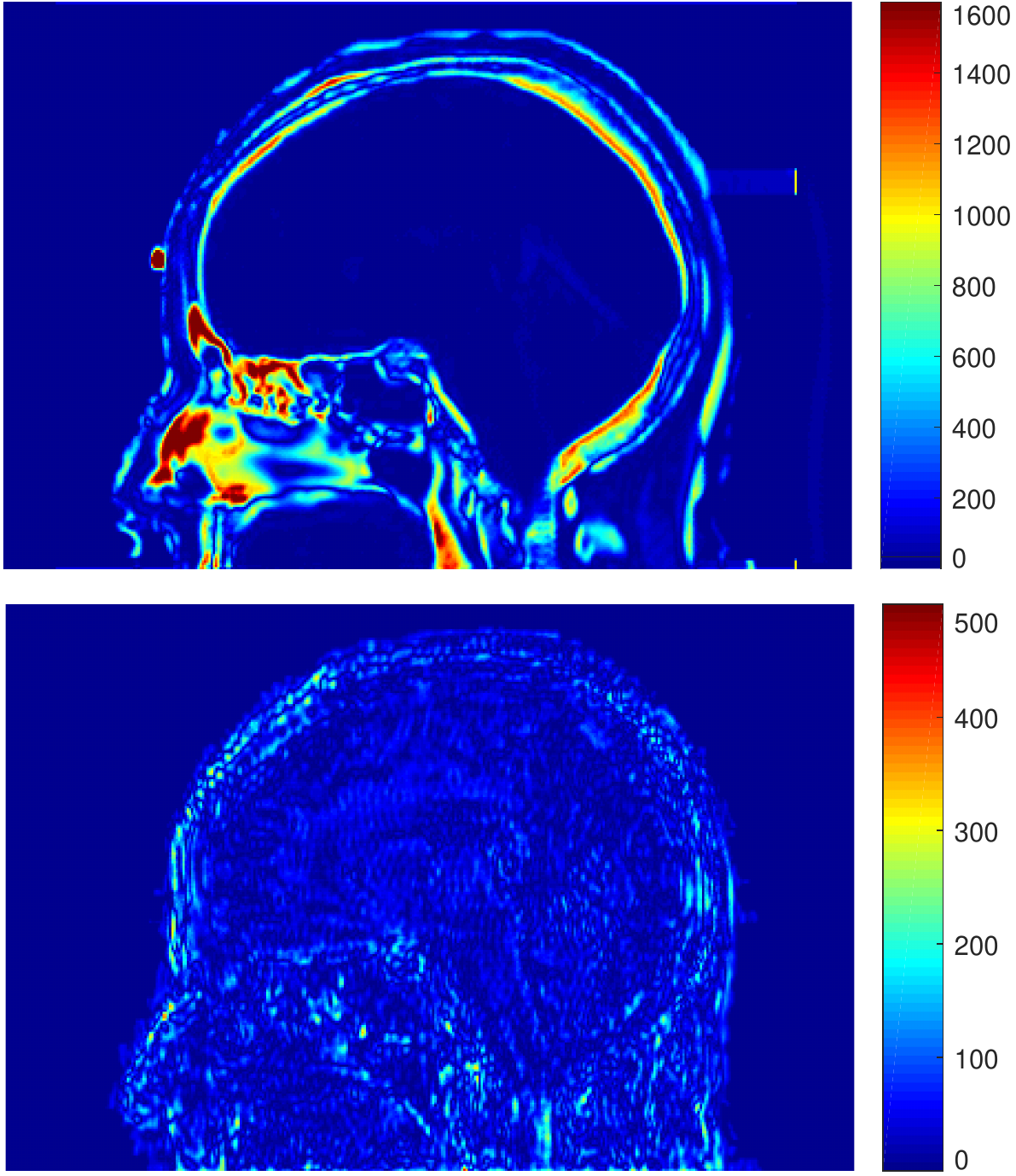}
		\end{minipage}
	}
	\caption{\small Visual example of a cycleGAN result. We show (a) ground-truth CT image and input MR image, (b) synthetic CT image and reconstructed MR image, and (c) the relative errors between the ground-truth/synthetic CT images (upper) and the input/reconstructed MR images (lower) .}
	\label{fig:motivation}
\end{figure*}

Although these methods can produce good synthetic images, they rely on
a large number of paired CT and MR images, which are hard to obtain in
practice, especially for specific MR tissue contrasts. To relax the
requirement of paired data, Wolterink et al.~\cite{wolterink2017} and
Chartsias et al.~\cite{chartsias2017} used a cycleGAN~\cite{zhu2017} for
MR-to-CT synthesis on unpaired data with promising results. They
used a CNN to learn the MR-to-CT mapping with the help of an
adversarial loss, which forces synthetic CT images to be
indistinguishable from real CT images. To ensure the synthetic CT
image correctly corresponds to an input MR image, another CNN is
utilized to map synthetic CT back to the MR domain and the
reconstructed image should be identical to the input MR image
(i.e., cycle-consistency loss).

However, due to a lack of direct constraints between the synthetic and
input images, the cycleGAN cannot guarantee structural consistency
between these two images.
As shown in Fig.~\ref{fig:motivation}, the reconstructed MR image is almost identical to the input MR image, indicating the cycle consistency is well kept, but the synthetic CT image is quite different from the ground-truth, especially for the skull region, which illustrates that the structure of the synthetic CT image is not consistent with that of the input MR image.
To overcome this, Zhang et al.~\cite{zhang2018} trained two auxiliary CNNs respectively for segmenting MR and CT images and also defined a loss to force the segmentation of the synthetic image to be the same as the ground-truth segmentation of the input image. This requires a training dataset with ground-truth segmentations of MR and CT images, which further complicates the training data requirements.

In this work, we propose a structure-constrained cycleGAN to constrain structural consistency without requiring ground-truth segmentations. By using the modality independent neighborhood descriptor~\cite{heinrich2012}, we define a structure-consistency loss enforcing the extracted features in the synthetic image to be voxel-wise close to the ones extracted in
the input image. Additionally, we use a position-based selection strategy for selecting training images instead of a completely random selection scheme. Experimental results on synthesizing CT images from brain MR images show that our method achieves significantly better results compared to a conventional cycleGAN with various metrics, and approximates the cycleGAN trained with paired data.

\begin{figure}[!tp]
	\setlength{\abovecaptionskip}{2mm}
	\setlength{\belowcaptionskip}{-4mm}
	\centering
	\includegraphics[width=9cm]{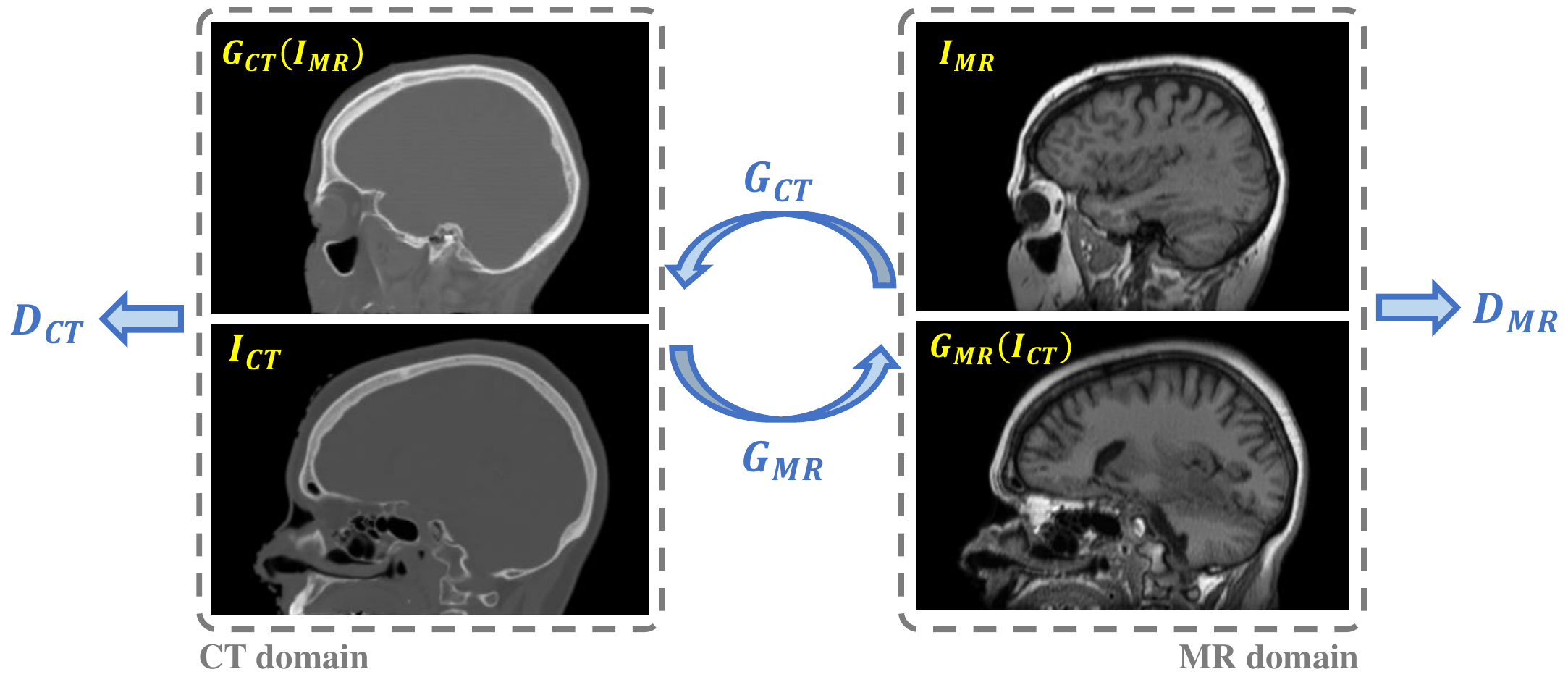}
	\caption{\small Illustration of our proposed structure-constrained cycleGAN. Two generators (i.e., $\GCT$ and $\GMR$) learn cross-domain mappings between CT and MR domains. The training of these mappings is supervised by adversarial, cycle-consistency, and structure-consistency losses.}
	\label{fig:sturcture}
\end{figure}

\section{Method}
In this section, we introduce our proposed structure-constrained cycleGAN. As shown in Fig.~\ref{fig:sturcture}, our method
contains two generators $\GCT$ and $\GMR$, which provide the MR-to-CT and CT-to-MR mappings, respectively. In addition, discriminator $\DCT$ is
used to distinguish between real and synthetic CT images, and
discriminator $\DMR$ is for MR images. Our training loss includes three
types of terms: an adversarial loss~\cite{Goodfellow2014} for matching the distribution of synthetic images to target CT or MR domain; a cycle-consistency loss~\cite{zhu2017} to prevent generators from producing synthetic images that are irrelevant to the inputs; and a structure-consistency loss to constrain structural consistency between input and synthetic images.
\subsection{Adversarial loss}
The adversarial loss~\cite{Goodfellow2014} is applied to both
generators. For the generator $\GCT$ and its discriminator $\DCT$, the
adversarial loss is defined as
\begin{equation}
\LGAN (\GCT, \DCT) = \DCT (\GCT (\IMR))^2 + \left( 1- \DCT
(\ICT) \right) ^2 \ ,
\end{equation}
where $\ICT$ and $\IMR$ denote the unpaired input CT and MR images.
During the training phase, $\GCT$ tries to generate a synthetic CT image
$\GCT (\IMR)$ close to a real CT image, i.e., $\max_{\GCT}
\LGAN (\GCT, \DCT)$, while $\DCT$ is to distinguish
between a synthetic CT image $\GCT (\IMR)$ and a real image $\ICT$,
i.e., $\min_{\DCT} \LGAN (\GCT, \DCT)$.  Similarly,
the adversarial loss for $\GMR$ and $\DMR$ is defined as 
\begin{equation}
\LGAN (\GMR, \DMR) = \DMR (\GMR (\ICT)) ^ 2 + \left( 1-
\DMR (\IMR) \right) ^2 \ .
\end{equation}

\subsection{Cycle-consistency loss}
To prevent the generators from producing synthetic images that are irrelevant to the
inputs, a cycle-consistency loss~\cite{zhu2017} is utilized for
$\GCT$ and $\GMR$ forcing the reconstructed images $\GCT
\left( \GMR (\ICT) \right)$ and $\GMR \left( \GCT (\IMR) \right)$ to
be identical to their inputs $\ICT$ and $\IMR$. This loss is written as
\begin{equation}
\begin{aligned}
	\Lc (\GCT, \GMR) & = \Vert \GCT \left( \GMR (\ICT) \right) - \ICT
	\Vert_1 \\ & \quad + \Vert \GMR \left( \GCT (\IMR) \right) - \IMR
	\Vert_1 \ .
\end{aligned}
\end{equation}
\begin{figure*}[!tp]
	\setlength{\abovecaptionskip}{2mm}
	\setlength{\belowcaptionskip}{-1mm}
	\centering
	\subfloat[]{
		\label{fig:MIND_a}
		\begin{minipage}[t]{0.23\textwidth}
			\centering
			\includegraphics[width=2.9cm]{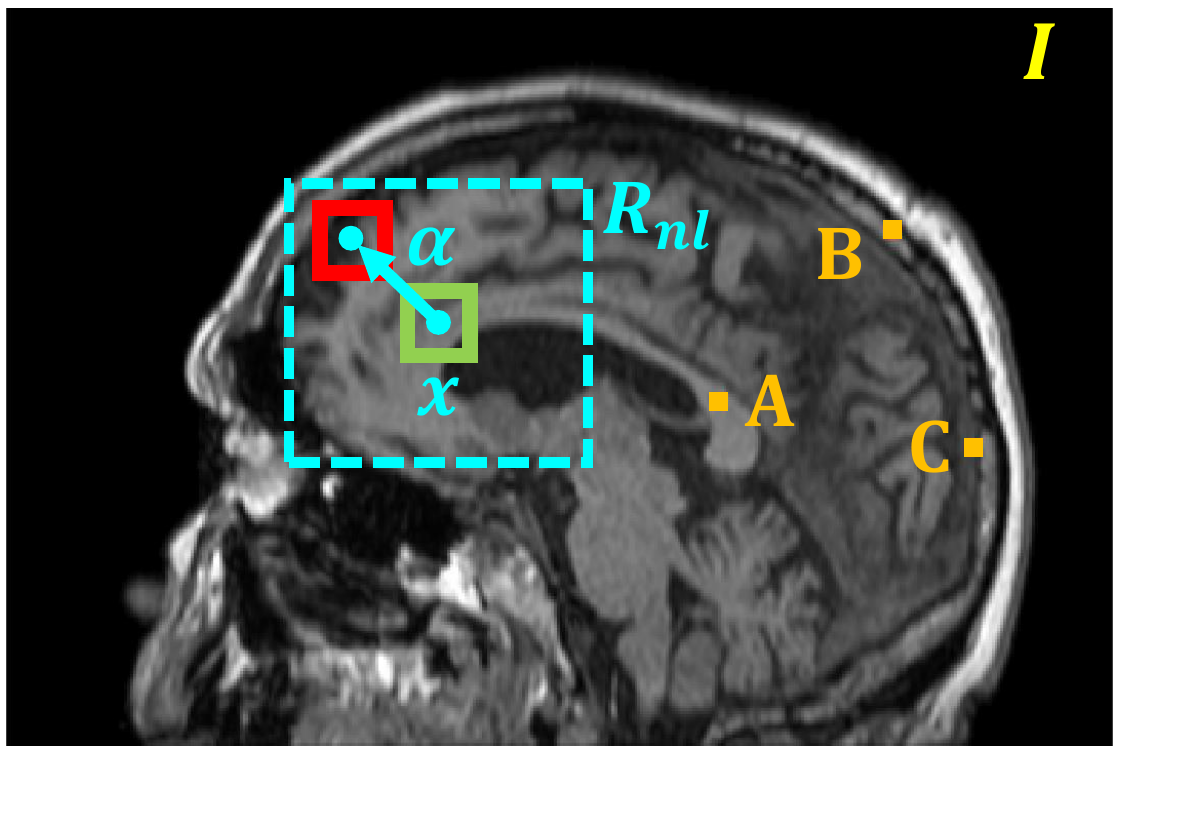}
		\end{minipage}
	}
	\subfloat[]{
		\label{fig:MIND_b}
		\begin{minipage}[t]{0.23\textwidth}
			\centering
			\includegraphics[width=2.9cm]{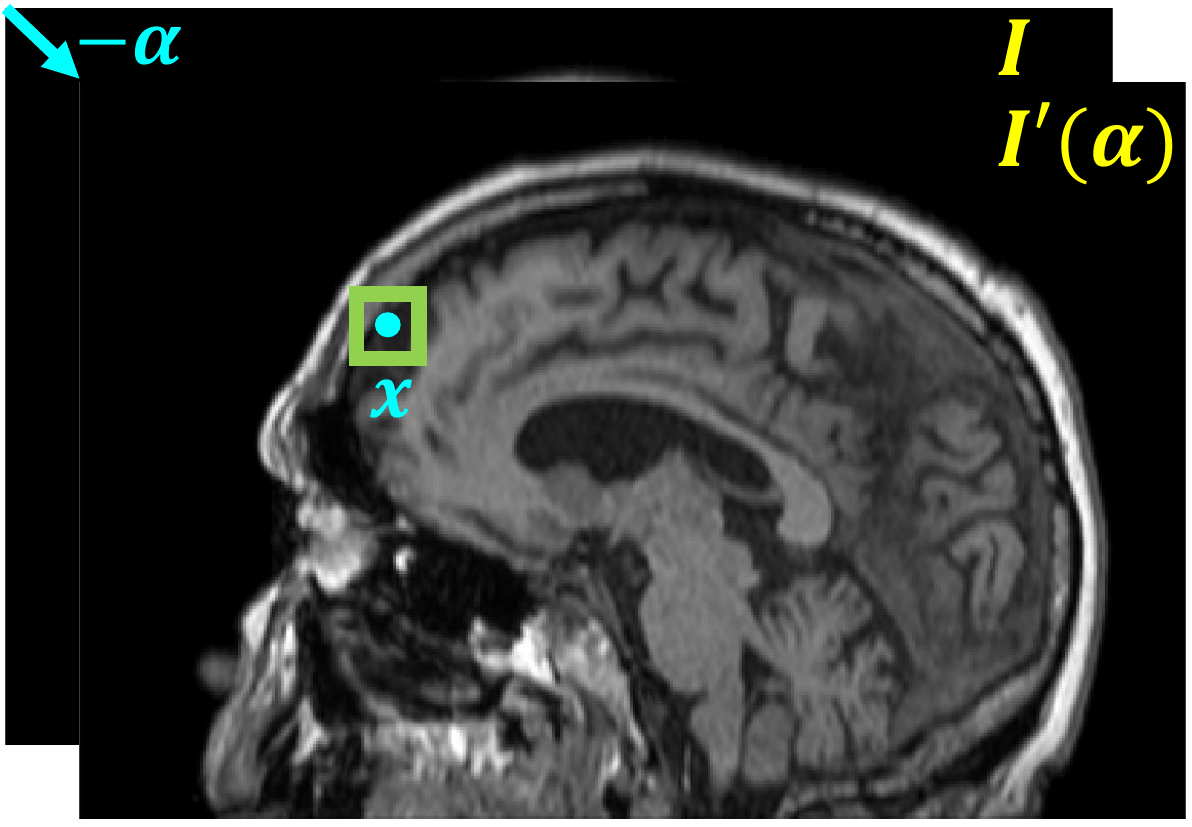}
		\end{minipage}
	}
	\subfloat[]{
		\label{fig:MIND_c}
		\begin{minipage}[t]{0.23\textwidth}
			\centering
			\includegraphics[width=2.9cm]{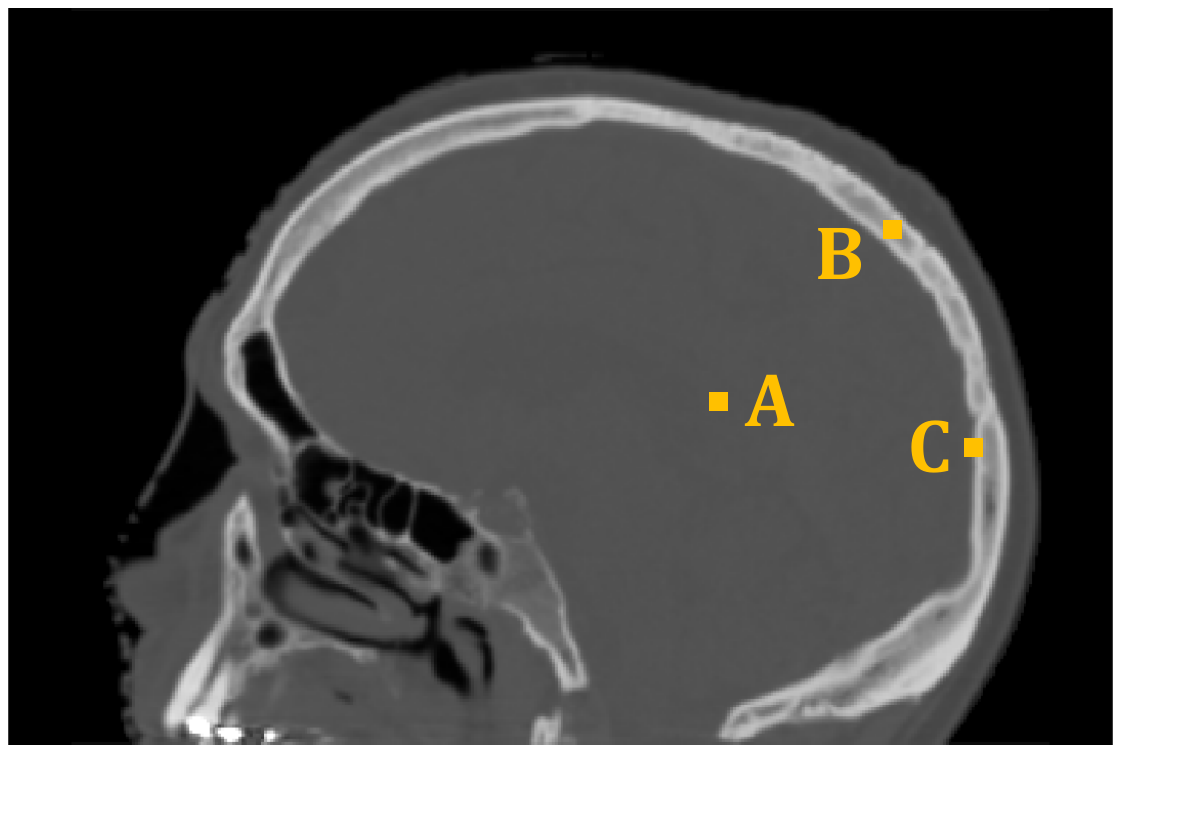}
		\end{minipage}
	}
	\subfloat[]{
		\label{fig:MIND_d}
		\begin{minipage}[t]{0.23\textwidth}
			\centering
			\includegraphics[width=2.9cm]{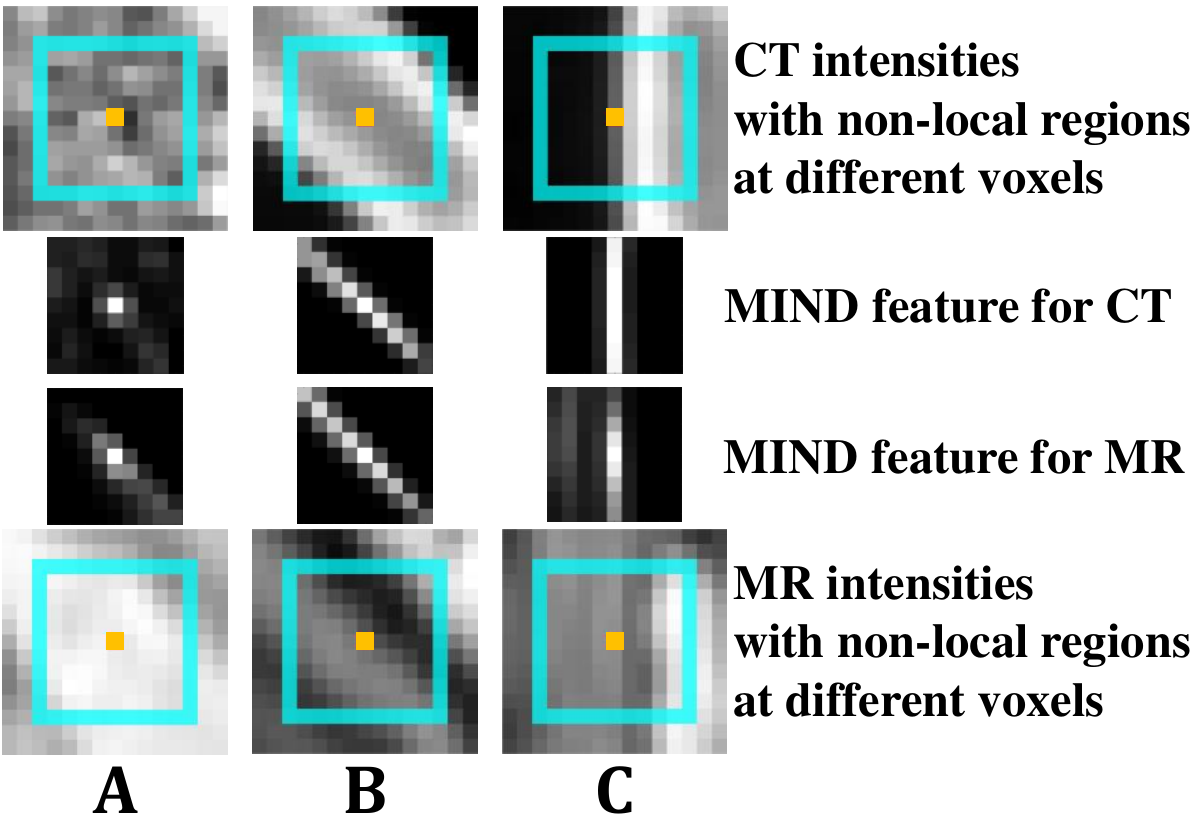}
		\end{minipage}
	}
	\caption{\small Illustration of the MIND feature. (a) To extract the MIND feature at $x$, a patch around $x+\alpha$ is compared with a patch around $x$ for each $x+\alpha \in R_{nl}$; (b) comparison between  $x$ and $x+\alpha$ of $I$ in (a) equals a comparison of $I$ and $I'(\alpha)$ at $x$; (c) the CT image paired with MR image in (a); (d) visual examples of MIND features extracted at voxels $A,B,C$ within paired MR and CT images in (a) and (c).}
	\label{fig:MIND}
\end{figure*}

\subsection{Structure-consistency loss}
Since the cycle-consistency loss does not necessarily ensure structural consistency (as discussed in Sec.~\ref{sec:introduction}), our method uses an extra structure-consistency loss between the synthetic and input images.
However, as these two images are respectively in MR and CT domains, we first map these images into a common feature domain by using a modal-independent structural feature, and then the structural consistency between the synthetic and input images is measured in this feature domain.
In this work, we use the modality independent neighborhood descriptor~(MIND)~\cite{heinrich2012} as the structural feature. MIND is defined using a non-local patch-based self-similarity and depends on image local structure instead of intensity values. It has been previously applied to MR/CT image registration as a similarity metric.
Figure~\ref{fig:MIND}(d) shows visual examples of MIND features extracted at different voxels in MR and CT images.
In the following paragraphs, we introduce the MIND feature and our structure-consistency loss in detail.

The MIND feature extracts distinctive image structure by comparing each patch with all its neighbors in a non-local region.
As shown in Fig.~\ref{fig:MIND}(a), for voxel $x$ in image $I$, the MIND feature $F_{x}$ is an $|R_{nl}|$-length vector, where $R_{nl}$ denotes a non-local region around voxel $x$, and each component $F_{x} ^{(\alpha)}$ for a voxel $x + \alpha \in R_{nl}$ is defined as
\begin{equation}
	F_{x} ^{(\alpha)} (I) = \frac{1}{Z} \exp \left(- \frac{D_{\mathcal{P}} (I, x, x+\alpha)}{V(I,x)} \right) \ ,
\end{equation}
where $Z$ is a normalization constant so that the maximal component of $F_{x}$ is 1.
$D_{\mathcal{P}} (I, x, x+\alpha)$ denotes the $L_2$ distance between two image patches $\mathcal{P}$ respectively centered at voxel $x$ and voxel $x+\alpha$ in image $I$, and $V(I, x)$ is an estimation of local variance at voxel $x$, which can be written as
\begin{eqnarray}
\label{eqn:Dp}
D_{\mathcal{P}} (I, x, x+\alpha) & = & \sum_{p \in \mathcal{P}} \left( I(x+p) - I(x+\alpha + p) \right) ^ 2 \ , \\
V(I, x) & = & \frac{1}{4} \sum_{n \in \mathcal{N} } D_{\mathcal{P}} (I, x, x+n) \ ,
\end{eqnarray}
where $\mathcal{N}$ is the 4-neighborhood of voxel $x$.

It is difficult to directly compute the operation $D_{\mathcal{P}}$ and its gradient using Eqn.~\ref{eqn:Dp} in a deep network.
Instead, as shown in Fig.~\ref{fig:MIND}(b), $D_{\mathcal{P}}$ can be equivalently computed
by using a convolutional operation as
\begin{equation}
\label{equ:gpu_implement}
	D_{\mathcal{P}} (I, x, x+ \alpha) = C * (I - I'(\alpha)) ^ 2 \ ,
\end{equation}
where $C$ is an all-one kernel of the same size as patch $\mathcal{P}$, and $I'(\alpha)$ denotes $I$ translated by $\alpha$.
By doing this, the structural feature can be extracted via several simple operations and the gradients of these operations can be easily computed.

Based on the MIND feature introduced above, the structure-consistency loss in our method is defined to enforce the extracted MIND features in the synthetic images $\GCT (\IMR)$ or $\GMR (\ICT)$ to be voxel-wise close to the ones extracted in their inputs $\IMR$ or $\ICT$, which can be written as
\begin{equation}
\begin{aligned}
	\Lstructure (\GCT, \GMR) = & \frac{1}{\NMR |R_{nl}|} \sum_{x} \Vert
	F_x (\GCT (\IMR)) - F_x (\IMR) \Vert _1 \\
	& + \frac{1}{\NCT |R_{nl}|} \sum_{x} \Vert
	F_x (\GMR (\ICT)) - F_x (\ICT) \Vert _1 \ , \\
\end{aligned}
\end{equation}
where $\NMR$ and $\NCT$ respectively denote the number of voxels in
input images $\IMR$ and $\ICT$, and $\Vert \cdot \Vert_1$ is the $L_1$ norm.
In this work, we use a $9 \times 9$ non-local region and a $7 \times
7$ patch for computing structure-consistency loss. 
Furthermore, instead of an all-one kernel $C$, we utilize a Gaussian kernel $C_{\sigma}$ with standard deviation $\sigma = 2$ to reweight the importance of voxels within patch $\mathcal{P}$ in Eqn.~\ref{equ:gpu_implement}.
In preliminary experiments, we tried different non-local regions, patch sizes, and $\sigma$ values, but did not observe improved performance.

\subsection{Training loss}
Given the definitions of adversarial, cycle-consistency, and
structure-consistency losses above, the training loss of our proposed
method is defined as:
\begin{equation}
\begin{aligned}
\mathcal{L} (\GCT, \GMR, \DCT, & \DMR) = \LGAN (\GCT, \DCT) + \LGAN
(\GMR, \DMR)\\
& + \lambda_1 \Lc (\GCT, \GMR) + \lambda_2 \Lstructure (\GCT, \GMR) \ ,\\
\end{aligned}
\end{equation}
where $\lambda_1$ and $\lambda_2$ control the relative importance of
the loss terms. During training, $\lambda_1$ is set to 10
as per~\cite{wolterink2017, zhu2017} and $\lambda_2$ is set to 5.
To optimize $\mathcal{L}$, we alternatively update $D_{\mbox{\tiny{MR/CT}}}$ (with $G_{\mbox{\tiny{MR/CT}}}$ fixed) and $G_{\mbox{\tiny{MR/CT}}}$ (with $D_{\mbox{\tiny{MR/CT}}}$ fixed).

\subsection{Network structure}
Our method is composed of four trainable neural networks, i.e.,
two generators, $\GCT$ and $\GMR$, and two discriminators, $\DCT$ and $\DMR$, and
we use the same network structures as~\cite{zhu2017,
wolterink2017} in this work. That is, two generators, $\GCT$ and $\GMR$, are
2D fully convolutional networks (FCNs) with two stride-2 convolutional
layers, nine residual blocks, and two fractionally-strided convolutional
layers with stride $\frac{1}{2}$. The two discriminators, $\DCT$ and $\DMR$, are 2D FCNs consisting of five convolutional layers to classify whether $70 \times 70$ overlapping image patches are real or synthetic. For further details, please refer to~\cite{zhu2017}.

\subsection{Position-based selection strategy}
\label{sec:input_selection_strategy}
Although our input MR and CT slices are unpaired, we can get the positions of
their slices within the volumes. Slices in the middle of the volume necessarily have more brain tissue than peripheral slices. Thus, instead of feeding in slices at extremely different positions of the brain, e.g., a peripheral CT slice and a medial MR slice, we input training slices at similar positions; this is referred to as a position-based selection~(PBS) strategy.
That is, the MR and CT slices are linearly aligned considering their respective numbers of slices within the volumes, and given the $i$-th MR slice in its volume, the index $T(i)$ of corresponding CT slice selected by our method is determined by
\begin{equation}
	T(i) = \left\{
	\begin{array}{rcl}
	\left[ i \cdot \frac{\KCT - 1}{\KMR - 1} \right] + \ m \ ,    &  \quad  & \mbox{if} \ 5 \leq \left[ i \cdot \frac{\KCT - 1}{\KMR - 1} \right]  <  \KCT - 5 ,  \\
	\left[ i \cdot \frac{\KCT - 1}{\KMR - 1} \right]  \ ,     &  \quad   & \mbox{otherwise} ,
	\end{array} \right .
\end{equation}
where $\KMR$ and $\KCT$ respectively denote the number of slices in unpaired MR and CT volumes. $[ \cdot ]$ denotes the rounding function, and $m$ is a random integer within the range of $[-5,5]$.
This strategy forces the discriminators to be stronger at distinguishing synthetic images from real ones, thus avoiding mode collapse. This in turn forces our generators to be better in order to \textit{trick} our discriminators. 
We evaluate this position-based selection strategy in Sec.~\ref{sec:experiment}.

\section{Experiments}
\label{sec:experiment}
\subsection{Data set}
The MR and CT volumes are respectively obtained using a Siemens
Magnetom Espree 1.5T scanner~(Siemens Medical Solutions, Malvern, PA)
and a Philips Brilliance Big Bore scanner~(Philips Medical Systems,
Netherlands) under a routine clinical protocol for brain cancer patients. Geometric distortions in MR volumes are corrected using a 3D correction algorithm in the Siemens Syngo console workstation. All MR volumes are N4 corrected and normalized by aligning the white matter peak identified by fuzzy C-means.

The data set contains the brain MR and CT volumes of 45 patients, which were divided into a training set containing MR and CT volumes of 27 patients, a validation set of 3 patients for model and epoch selection, and a test set of 15 patients for performance evaluation.
As in~\cite{wolterink2017}, the experiments were performed on 2D sagittal image slices.
Each MR or CT volume contains about 270 sagittal images, which are resized and padded to $384 \times 256$ while maintaining the aspect ratio, and the intensity ranges are respectively $[-1000,3500]$ HU for CT and $[0,3500]$ for MR.
To augment the training set, each image is padded to $400 \times 284$ and then randomly cropped to $384 \times 256$ as training samples.
\begin{figure*}[!tp]
	\setlength{\abovecaptionskip}{2mm}
	\setlength{\belowcaptionskip}{-4mm}
	\centering
	\subfloat[MAE]{
		\label{fig:box_a}
		\begin{minipage}[t]{0.23\textwidth}
			\centering
			\includegraphics[width=2.8cm]{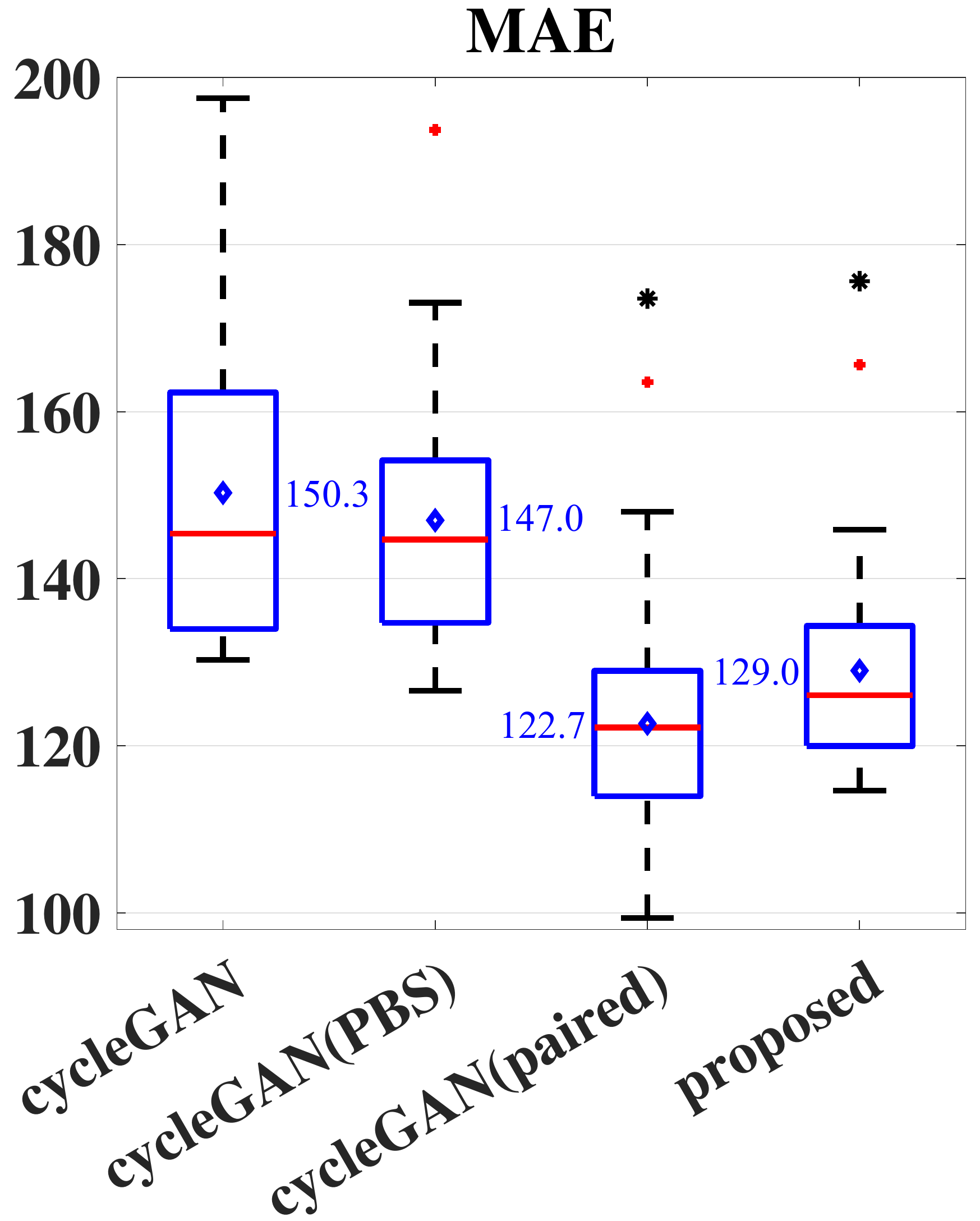}
		\end{minipage}
	}
	\subfloat[PSNR]{
		\label{fig:box_b}
		\begin{minipage}[t]{0.23\textwidth}
			\centering
			\includegraphics[width=2.8cm]{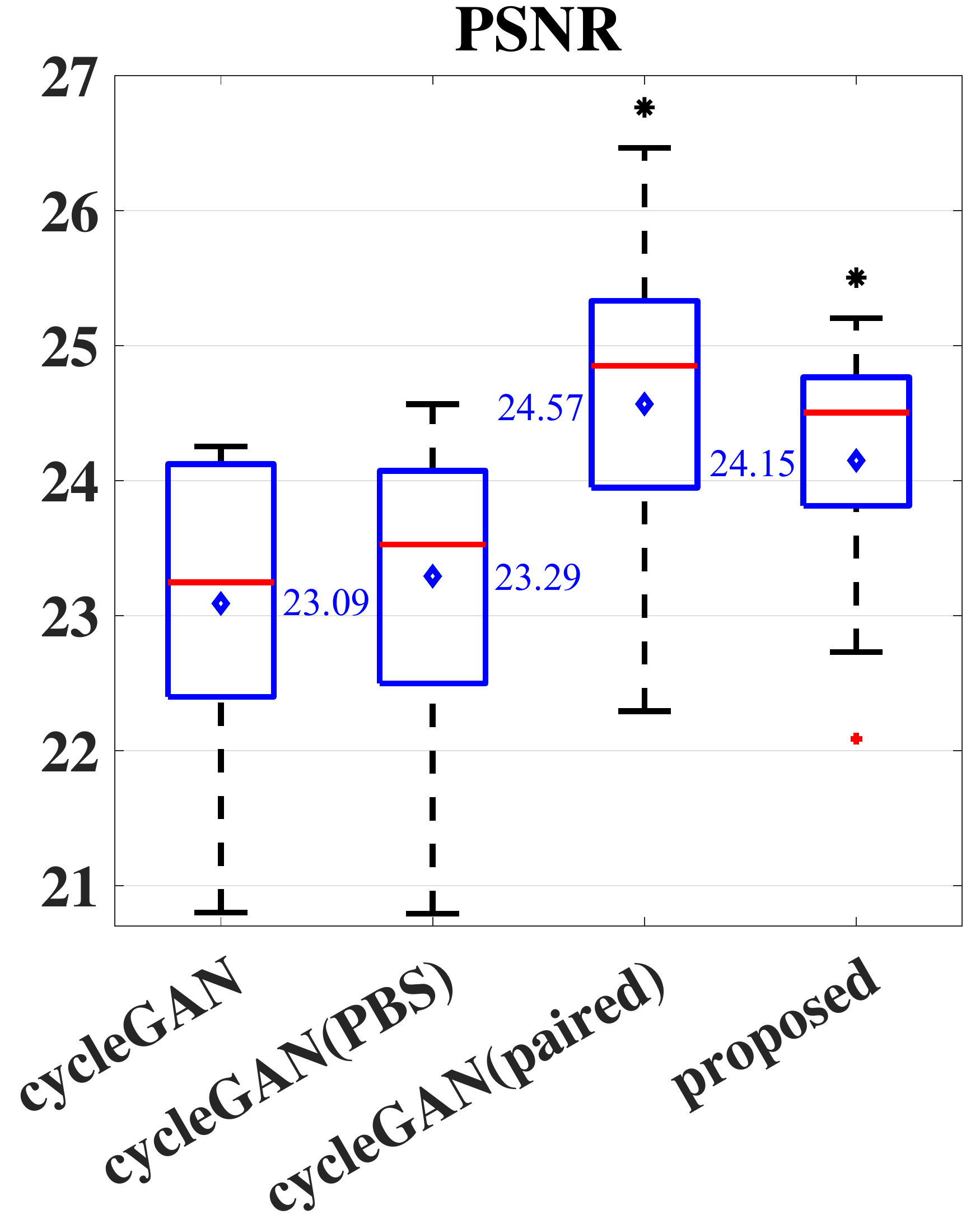}
		\end{minipage}
	}
	\subfloat[SSIM]{
		\label{fig:box_c}
		\begin{minipage}[t]{0.23\textwidth}
			\centering
			\includegraphics[width=2.8cm]{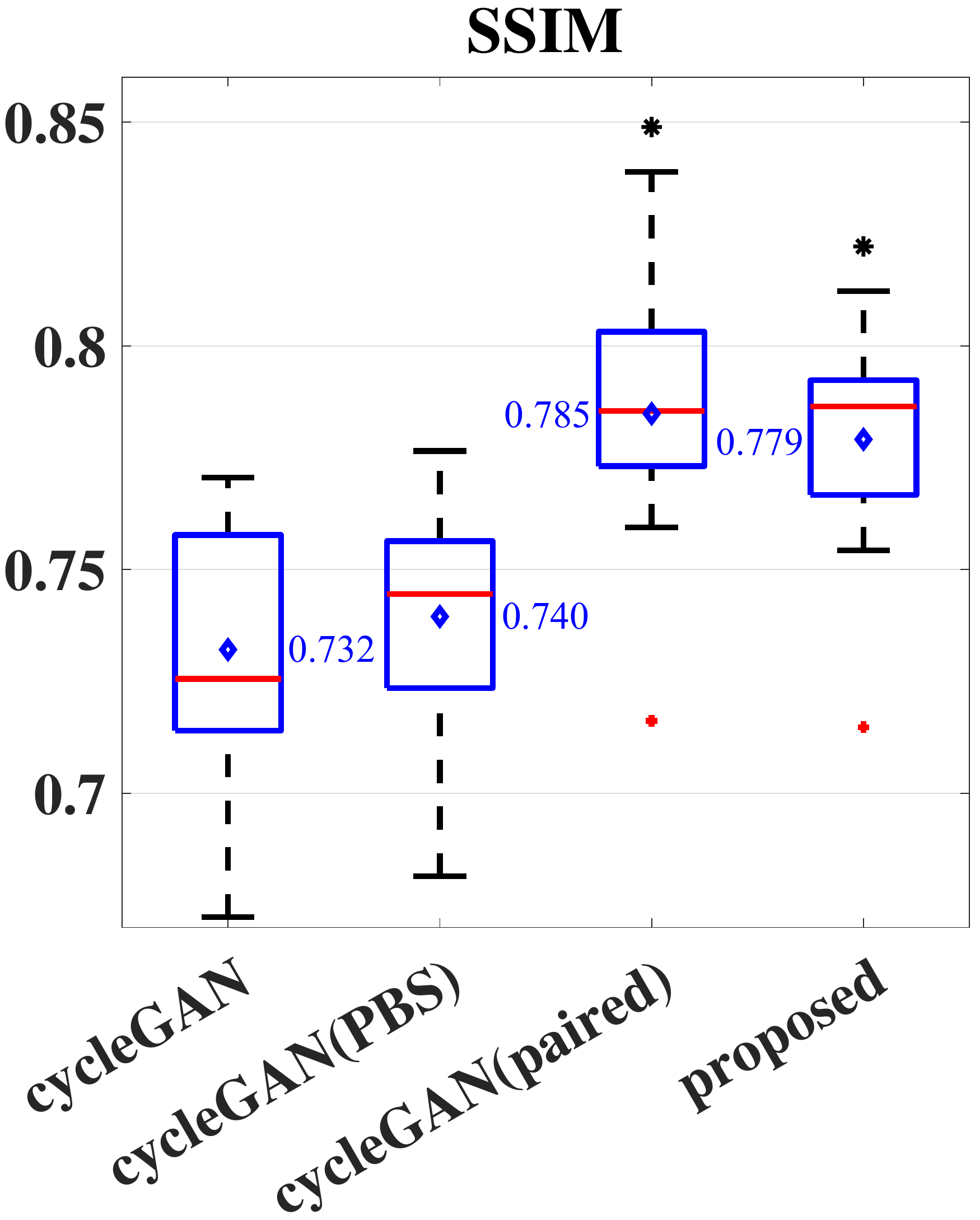}
		\end{minipage}
	}
	\subfloat[SSIM(HG)]{
		\label{fig:box_d}
		\begin{minipage}[t]{0.23\textwidth}
			\centering
			\includegraphics[width=2.8cm]{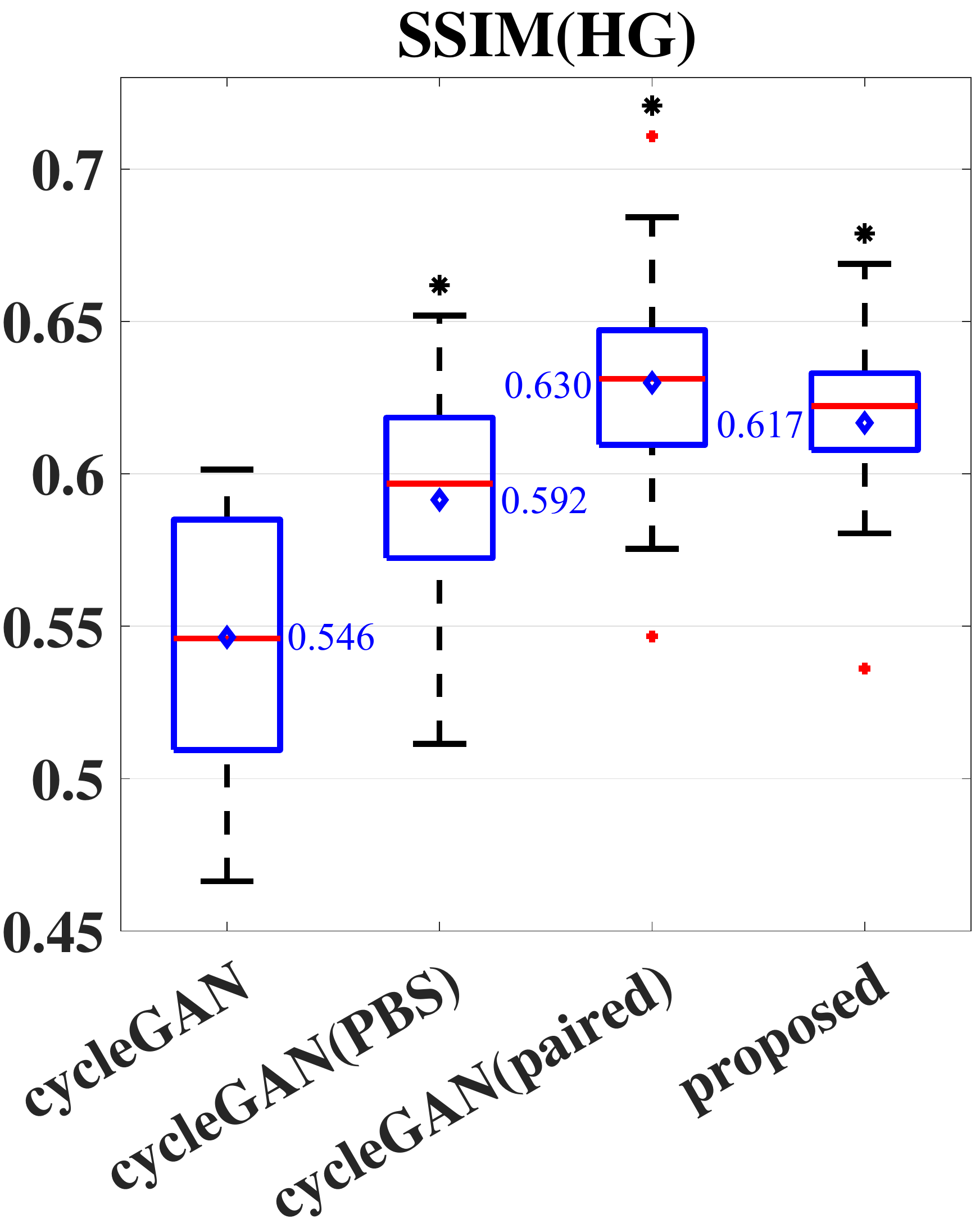}
		\end{minipage}
	}
	\caption{\small Comparison of different methods on synthesizing CT images in boxplots, where the diamond and number in blue denote the respective mean and $\ast$ denotes $p<0.001$ compared to the conventional cycleGAN using a paired sample t-test. }
	\label{fig:boxplot}
\end{figure*}

\subsection{Experimental results}
We compare the proposed method to the conventional
cycleGAN~\cite{zhu2017,wolterink2017} (denoted as ``cycleGAN'') and a
cycleGAN trained with paired data (denoted as ``cycleGAN
(paired)''), which represents the best that a cycleGAN can achieve.
To evaluate the position-based selection strategy
in Sec.~\ref{sec:input_selection_strategy}, a cycleGAN using this
strategy during training, denoted as ``cycleGAN (PBS)'', is also included in comparison.
As in \cite{zhu2017, wolterink2017}, the learning rate is set to 0.0002 for all compared methods.

To quantitatively compare these methods, we use mean absolute
error~(MAE), peak signal-to-noise ratio~(PSNR), and structural
similarity~(SSIM) between the ground-truth CT volume and the synthetic
one, which are computed within the head region mask and averaged over
15 test subjects. Furthermore, SSIM over regions with high gradient magnitudes (denoted as ``SSIM(HG)'') is also computed to measure the quality of bone regions in synthetic images.
The maximum value in PSNR and the dynamic range in SSIM are set to 4500, as the range of our CT data is $[-1000,3500]$ HU.

\begin{figure*}[!tp]
	\setlength{\abovecaptionskip}{2mm}
	\setlength{\belowcaptionskip}{-2mm}
	\subfloat[]{
		\label{fig:visual_a}
		\begin{minipage}[t]{0.178\textwidth}
			\centering
			\includegraphics[width=2.35cm]{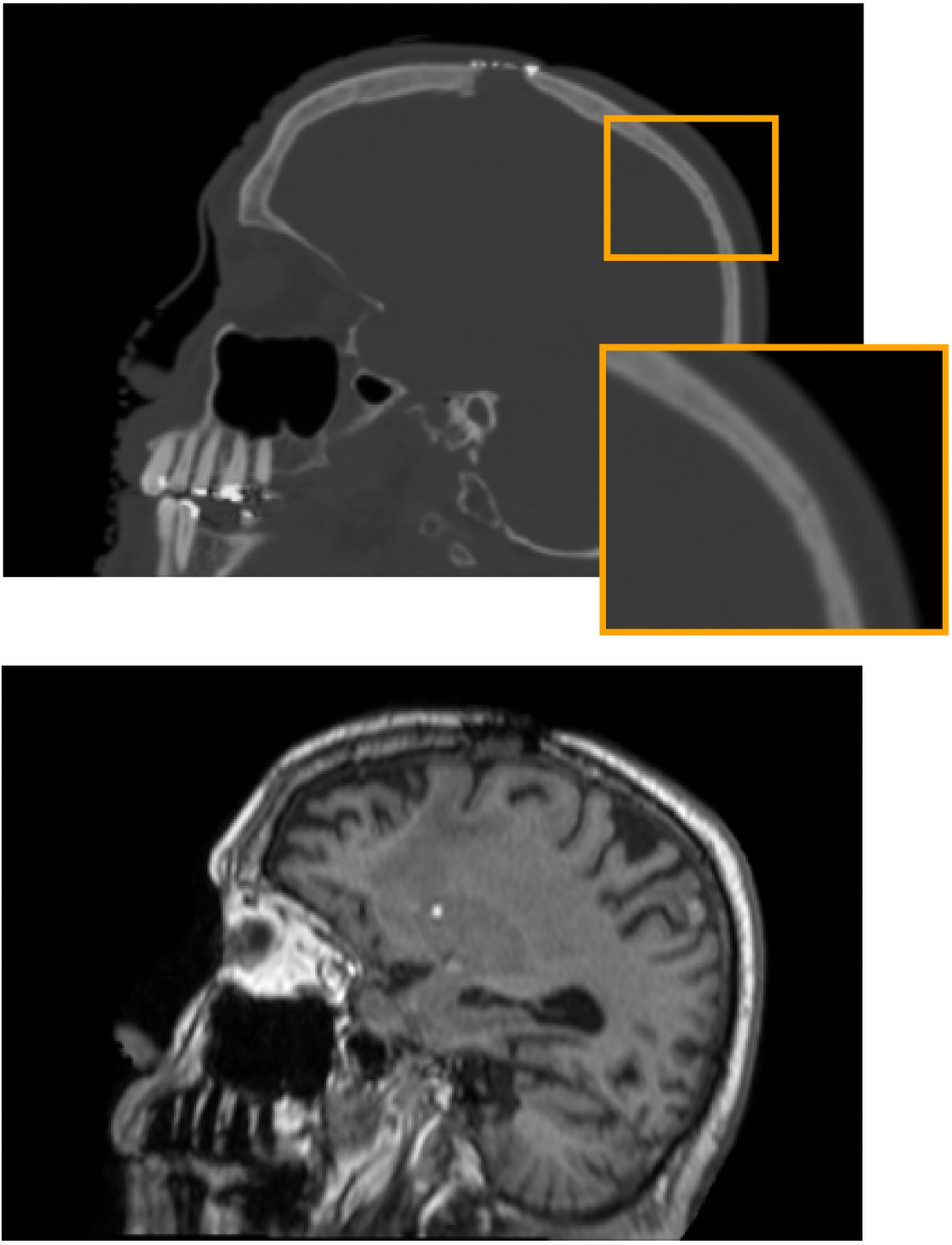}
		\end{minipage}
	}
	\subfloat[]{
		\label{fig:visual_b}
		\begin{minipage}[t]{0.178\textwidth}
			\centering
			\includegraphics[width=2.35cm]{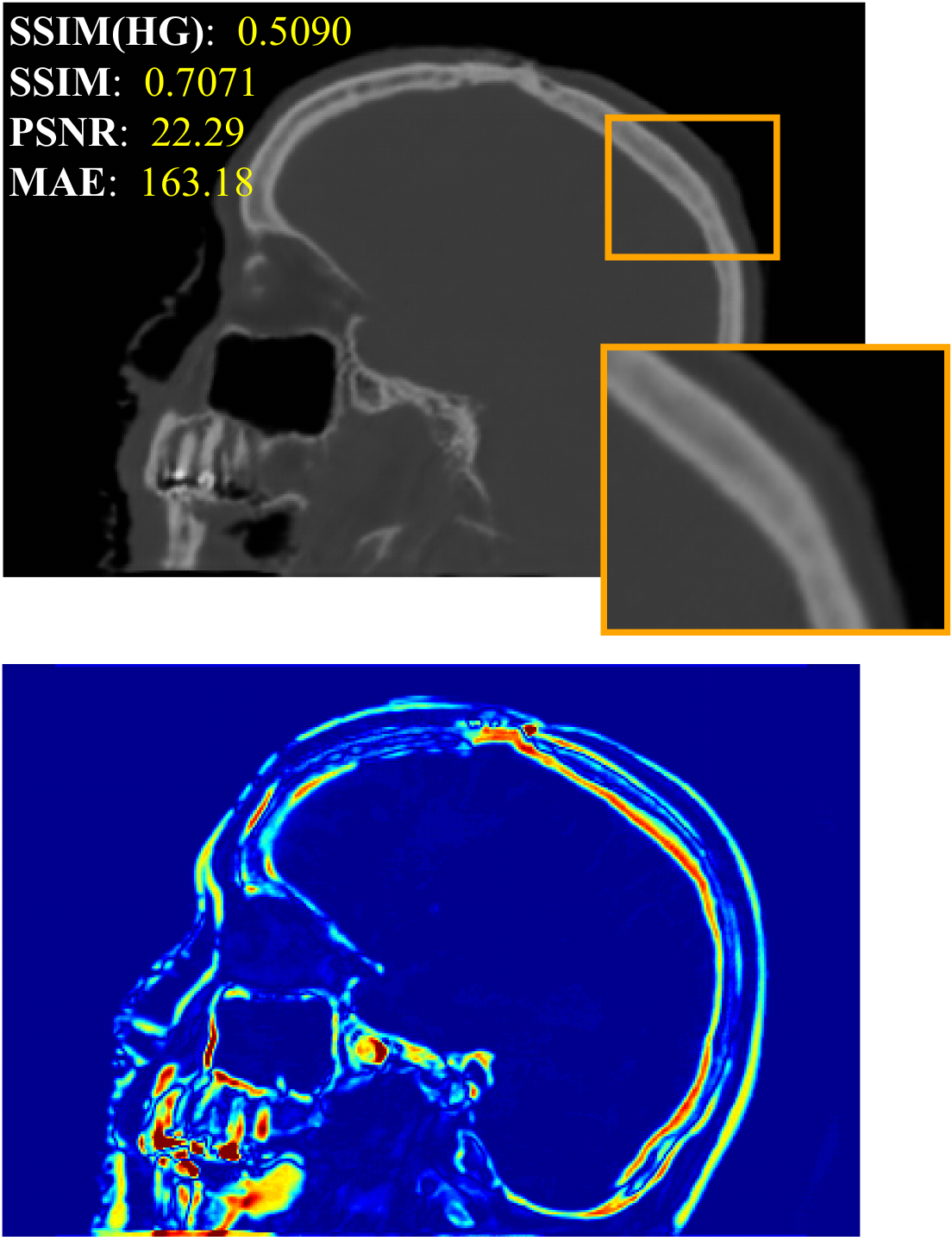}
		\end{minipage}
	}
	\subfloat[]{
		\label{fig:visual_c}
		\begin{minipage}[t]{0.178\textwidth}
			\centering
			\includegraphics[width=2.35cm]{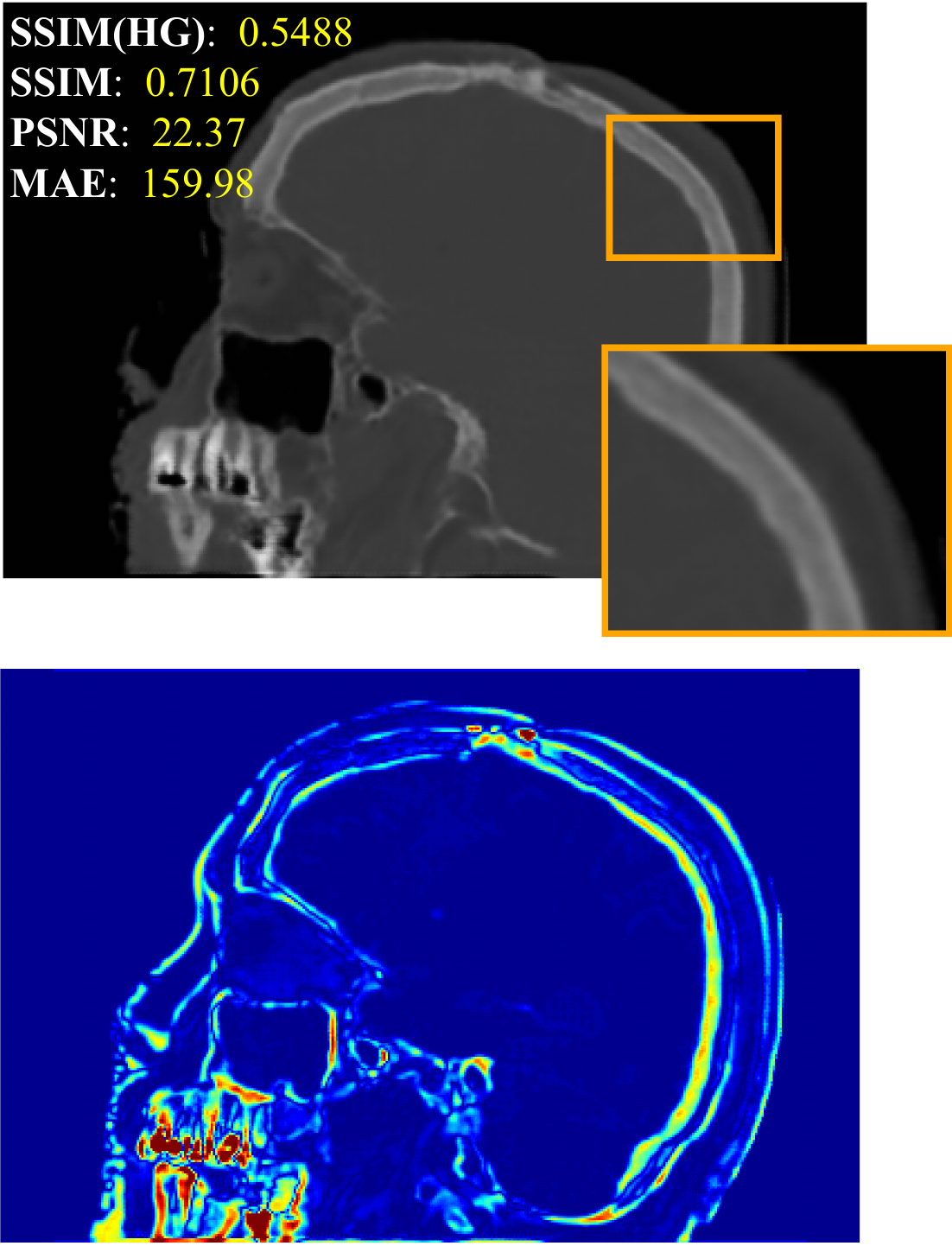}
		\end{minipage}
	}
	\subfloat[]{
		\label{fig:visual_d}
		\begin{minipage}[t]{0.178\textwidth}
			\centering
			\includegraphics[width=2.35cm]{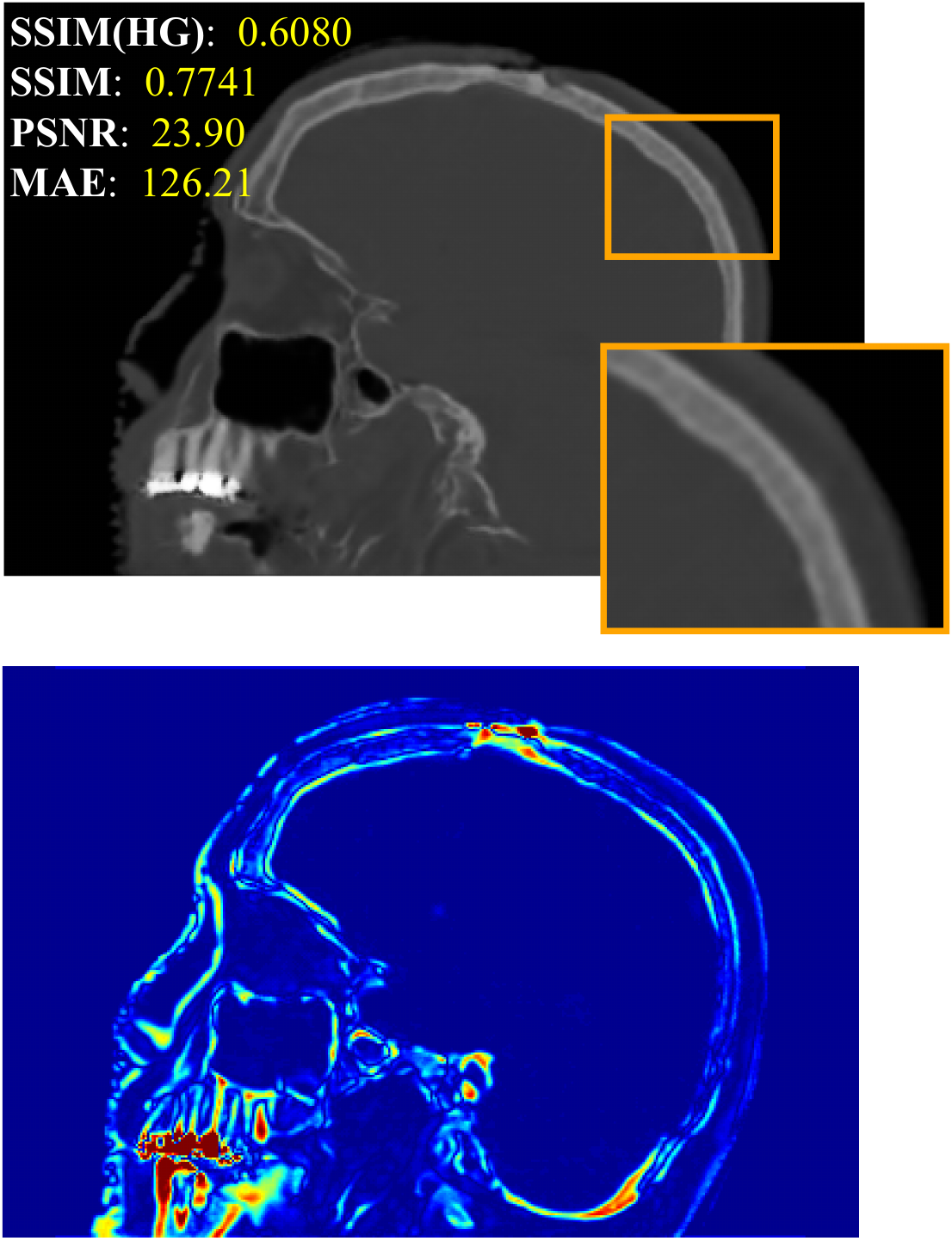}
		\end{minipage}
	}
	\subfloat[]{
		\label{fig:visual_e}
		\begin{minipage}[t]{0.178\textwidth}
			\centering
			\includegraphics[width=2.60cm]{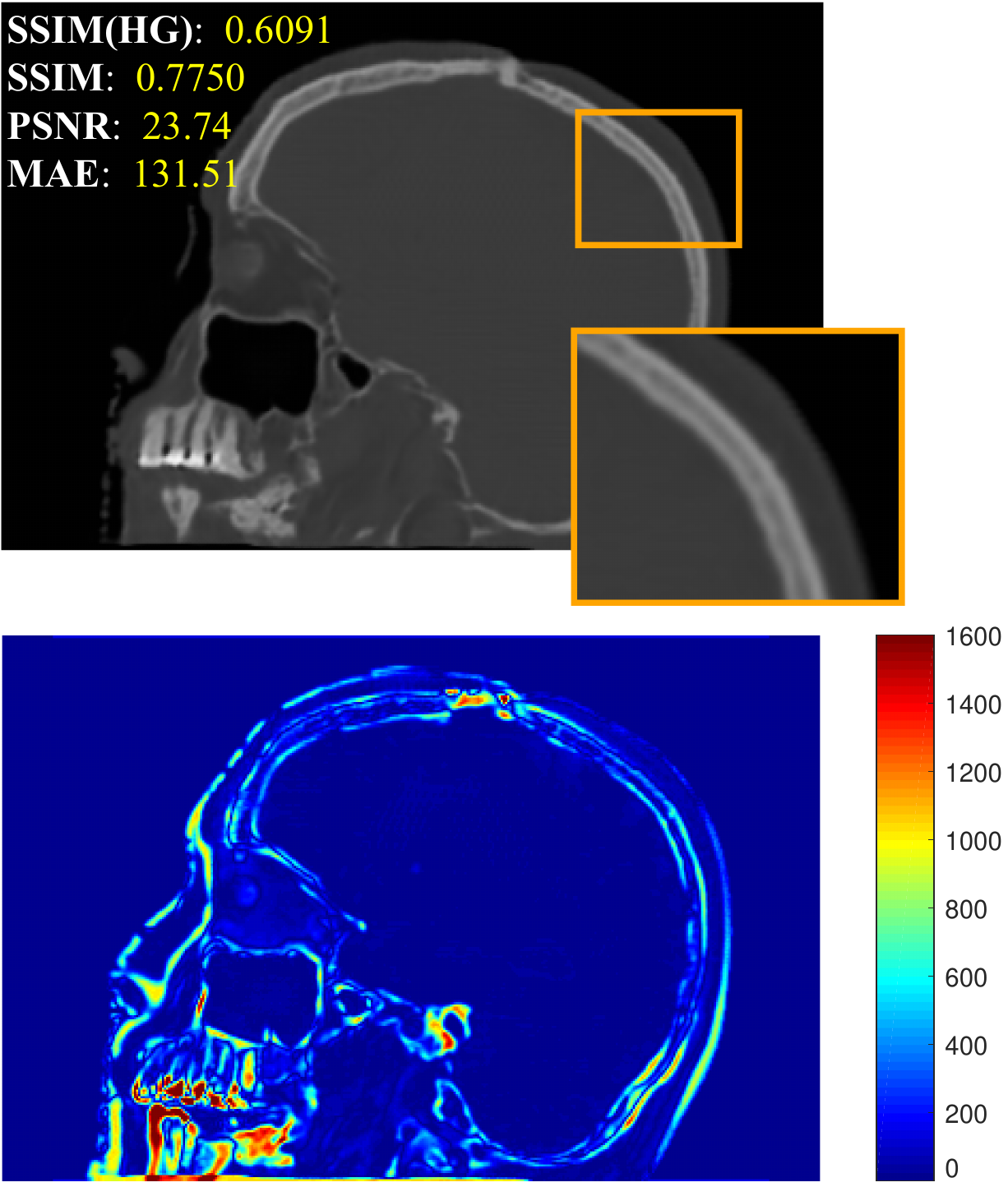}
		\end{minipage}
	}
	\caption{\small Visual comparison of synthetic CT images using different
		methods. For one test subject, we show (a)~the ground-truth CT
		image and input MR image; the synthetic CT image and its
		difference image (compared to ground-truth CT image) generated
		by (b)~cycleGAN, (c)~cycleGAN (PBS), (d)~cycleGAN (paired), and
		(e)~proposed method. The small text in each sub-image is the corresponding accuracy on this test subject.}
	\label{fig:visual}
\end{figure*}

As shown in Fig.~\ref{fig:boxplot}, our proposed method achieves significantly better performance than conventional cycleGAN in all the metrics~($p<0.001$) and produces similar results compared to the cycleGAN trained with paired data.
Compared to randomly selecting training slices at any position, our proposed position-based selection strategy produces significantly higher SSIM(HG) score~($p<0.001$) with marginal improvement in the other three metrics. Figure~\ref{fig:visual} shows visual examples of synthetic CT images by different methods from a test subject.

\section{Conclusion}
We propose a structure-constrained cycleGAN for brain MR-to-CT synthesis using unpaired data. Compared to the conventional cycleGAN~\cite{zhu2017,wolterink2017}, we define an extra structure-consistency loss based on the modality independent neighborhood descriptor to constrain structural consistency and also introduce a position-based selection strategy for selecting training images.
The experiments show that our method generates better synthetic CT
images than the conventional cycleGAN and produces results similar to a cycleGAN trained with paired data.

\subsubsection*{Acknowledgments.}
This work is supported by the NSFC (11622106, 11690011, 61721002) and the China Scholarship Council.

\bibliographystyle{splncs04}
\bibliography{mybib}

\end{document}